\documentclass[preprint,11pt]{elsarticle}

\usepackage[utf8]{inputenc}
\usepackage{amsfonts}
\usepackage{nicefrac}
\usepackage{mathtools}
\usepackage{microtype}
\usepackage{soul}
\usepackage{color}
\usepackage{amsmath}
\usepackage{graphicx}
\usepackage{subfigure}
\usepackage{algorithm}
\usepackage{url}
\usepackage{balance}
\usepackage{multirow}
\usepackage{balance}
\usepackage{bm}
\usepackage{dsfont}
\usepackage{algpseudocode}
\usepackage{rotating}

\usepackage{tikz}
\usetikzlibrary{shapes}

\DeclareMathOperator*{\argmax}{arg\,max}

\usepackage{accents}

\newcommand\munderbar[1]{%
  \underaccent{\bar}{#1}}

\makeatletter
\def\squarecorner#1{
    \pgf@x=\the\wd\pgfnodeparttextbox%
    \pgfmathsetlength\pgf@xc{\pgfkeysvalueof{/pgf/inner xsep}}%
    \advance\pgf@x by 2\pgf@xc%
    \pgfmathsetlength\pgf@xb{\pgfkeysvalueof{/pgf/minimum width}}%
    \ifdim\pgf@x<\pgf@xb%
        \pgf@x=\pgf@xb%
    \fi%
    \pgf@y=\ht\pgfnodeparttextbox%
    \advance\pgf@y by\dp\pgfnodeparttextbox%
    \pgfmathsetlength\pgf@yc{\pgfkeysvalueof{/pgf/inner ysep}}%
    \advance\pgf@y by 2\pgf@yc%
    \pgfmathsetlength\pgf@yb{\pgfkeysvalueof{/pgf/minimum height}}%
    \ifdim\pgf@y<\pgf@yb%
        \pgf@y=\pgf@yb%
    \fi%
    \ifdim\pgf@x<\pgf@y%
        \pgf@x=\pgf@y%
    \else
        \pgf@y=\pgf@x%
    \fi
    \pgf@x=#1.5\pgf@x%
    \advance\pgf@x by.5\wd\pgfnodeparttextbox%
    \pgfmathsetlength\pgf@xa{\pgfkeysvalueof{/pgf/outer xsep}}%
    \advance\pgf@x by#1\pgf@xa%
    \pgf@y=#1.5\pgf@y%
    \advance\pgf@y by-.5\dp\pgfnodeparttextbox%
    \advance\pgf@y by.5\ht\pgfnodeparttextbox%
    \pgfmathsetlength\pgf@ya{\pgfkeysvalueof{/pgf/outer ysep}}%
    \advance\pgf@y by#1\pgf@ya%
}
\makeatother

\pgfdeclareshape{square}{
    \savedanchor\northeast{\squarecorner{}}
    \savedanchor\southwest{\squarecorner{-}}

    \foreach \x in {east,west} \foreach \y in {north,mid,base,south} {
        \inheritanchor[from=rectangle]{\y\space\x}
    }
    \foreach \x in {east,west,north,mid,base,south,center,text} {
        \inheritanchor[from=rectangle]{\x}
    }
    \inheritanchorborder[from=rectangle]
    \inheritbackgroundpath[from=rectangle]
}

\journal{Applied Soft Computing}

\begin{document}

\begin{frontmatter}

\title{Bayesian Neural Architecture Search using A Training-Free Performance Metric}

\author[dlr,uma]{Andr\'es Camero\corref{cor1}}
\ead{andres.camerounzueta@dlr.de}
\cortext[cor1]{Correspondent author}

\author[liacs]{Hao Wang}
\ead{h.wang@liacs.leidenuniv.nl}

\author[uma]{Enrique Alba}
\ead{eat@lcc.uma.es}

\author[liacs]{Thomas B\"ack}
\ead{t.h.w.baeck@liacs.leidenuniv.nl}

\address[uma]{Universidad de M\'{a}laga, ITIS Software, Espa\~{n}a}
\address[dlr]{GermanAerospace Center (DLR), Remote Sensing Technology Institute (IMF), Germany}
\address[liacs]{Leiden University, LIACS, The Netherlands}

\begin{abstract}
Recurrent neural networks (RNNs) are a powerful approach for time series prediction. 
However, their performance is strongly affected by their architecture and hyperparameter settings.
The architecture optimization of RNNs is a time-consuming task, where the search space is typically a mixture of real, integer and categorical values. To allow for shrinking and expanding the size of the network, the representation of architectures often has a variable length. 
In this paper, we propose to tackle the architecture optimization problem with a variant of the Bayesian Optimization (BO) algorithm. To reduce the evaluation time of candidate architectures the Mean Absolute Error Random Sampling (MRS), a training-free method to estimate the network performance, is adopted as the objective function for BO. 
Also, we propose three fixed-length encoding schemes to cope with the variable-length architecture representation.
The result is a new perspective on accurate and efficient  design of RNNs, that we validate on three problems. Our findings show that 
1) the BO algorithm can explore different network architectures using the proposed encoding schemes and successfully designs well-performing architectures, and 
2) the optimization time is significantly reduced by using MRS, without compromising the performance as compared to the architectures obtained from the actual training procedure. 
\end{abstract}


\end{frontmatter}

\noindent
\textbf{Copyright notice:} This article has been accepted for publication in the journal \textit{Applied Soft Computing}. Cite as: Camero, A., Wang, H., Alba, E. and B\"ack, T., 2021. Bayesian neural architecture search using a training-free performance metric. Applied Soft Computing, p.107356. \url{https://doi.org/10.1016/j.asoc.2021.107356}


\section{Introduction}

With the advent of deep learning, deep neural networks (DNNs) have gained popularity, and they have been applied to a wide variety of problems~\cite{haykin2009neural,LeCun2015}.
When it comes to sequence modeling and prediction, Recurrent Neural Networks (RNNs) have proved to be the most suitable ones~\cite{LeCun2015}. 
Essentially, RNNs are feedforward networks with feedback connections. This feature allows them to capture long-term dependencies among the input variables.
Despite their good performance, they are very sensitive to their \emph{hyperparameter} configuration and hard to train~\cite{bengio1994learning,haykin2009neural,Ojha2017,Pascanu2013}.

Finding an appropriate hyperparameter setting has always been a difficult task. The conventional approach to tackle this problem is to do a trial/error exploration based on expert knowledge. In other words, a human expert defines an architecture, sets up a training method (usually a gradient descent-based algorithm), and performs the training of the network until some criterion is met. Lately, automatic methods based on optimization algorithms, e.g., grid search, evolutionary algorithms or Bayesian optimization (BO), have been proposed to \emph{replace} the human expert. However, due to the immense size and complexity of the search space, and the high computational cost of training a DNN, hyperparameter optimization still poses an open problem~\cite{haykin2009neural,Ojha2017}.

Different approaches have been proposed for improving the performance of  hyperparameter optimization, ranging from evolutionary approaches (a.k.a.~neuroevolution)~\cite{Ojha2017}, to techniques to speed up the evaluation of a DNN~\cite{Camero2018lowcost,Domhan2015}. 
Among these approaches, the \emph{Mean Absolute Error Random Sampling} (MRS)~\cite{Camero2018lowcost} poses a promising ``low-cost, training-free, rule of thumb'' alternative to evaluate the performance of an RNN, which drastically reduces the evaluation time.

In this study, we propose to tackle the architecture optimization problem with a hybrid approach. Specifically, we combine BO~\cite{jones1998efficient,movckus1975bayesian} for optimizing the architecture, MRS~\cite{Camero2018lowcost} for evaluating the performance of candidate architectures, and ADAM~\cite{kingma2014adam} (a gradient descent-based algorithm) truncated through time for training the \emph{final} architecture on a given problem. We benchmark our proposal on three problems (the sine wave, the filling level of 217 recycling bins in a metropolitan area, and the load demand forecast of an electricity company in Slovakia) and compare our results against the state-of-the-art. 

Therefore, the main contributions of this study are:\vspace{-0.2cm}
\begin{itemize}
    \item We define a method to optimize the architecture of an RNN based on BO and MRS that significantly reduces the time without compromising the performance (error),\vspace{-0.2cm}
    \item We introduce multiple alternatives to cope with the variable-length solution problem. Specifically, we study three encoding schemes and two penalty approaches (i.e., the infeasible representation and the constraint handling), and\vspace{-0.2cm}
    \item We propose a strategy to improve the performance of the surrogate model of BO for variable-length solutions based on the augmentation of the initial set of solutions, i.e., the \emph{warm-start}.
\end{itemize}

The remainder of this article is organized as follows: Section~\ref{sec:related} briefly reviews some of the most relevant works related to our proposal. Section~\ref{sec:approach} introduces our proposed approach. Section~\ref{sec:experiments} presents the experimental study, and Section~\ref{sec:conclusions} provides conclusions and future work.

\section{Related Work}
\label{sec:related}

In this Section, we summarize some of the most relevant works related to our proposal. First, we introduce the architecture optimization problem and some interesting proposals to tackle it in section~\ref{subsec:hyper-opt}. Second, we present the \emph{neuroevolution}, a research line for handling the problem (section~\ref{subsec:neuroevolution}). After briefly reviewing the Mean Absolute Error Random Sampling (MRS) method in section \ref{subsec:MRS} we finally introduce Bayesian Optimization in section~\ref{subsec:bayesopt}.

\subsection{Architecture Optimization}
\label{subsec:hyper-opt} 

The existing literature teaches us on the importance of optimizing the architecture of a deep neural network on a particular problem, including, for example, the type of activation functions, the number of hidden layers, and the number of units for each layer~\cite{Bergstra2011,Camero2018lion,Jozefowicz2015}. For DNNs, the architecture optimization task is usually faced by either manual exploration of the search space (that is usually guided by expert knowledge) or by automatic methods based on optimization algorithms, e.g., grid search, evolutionary algorithms or Bayesian optimization~\cite{Ojha2017}. 

The challenges here are three-fold: firstly, the search space is typically huge due to the fact that the number of the parameters increases in proportion to the number of layers. Secondly, the search space is usually a mixture of real (e.g., the weights), integer (e.g., the number of units in each layer) and categorical (e.g., the type of activation functions) values, resulting in a demanding optimization task: different types of parameters naturally require different approaches for handling them in optimization. Last, the architecture optimization falls into the family of expensive optimization problems as function evaluations in this case are highly time consuming (which is affected both by the size of training data and the depth of the architecture). In this paper, we shall denote the search space of architecture optimization as $\mathcal{H}$. The specification of $\mathcal{H}$ depends on the choice of encoding schemes of the architecture (see Section~\ref{subsec:encoding}).  

To tackle the mentioned issues, many alternatives have been explored, ranging from reducing the evaluation time of a configuration (e.g., early stopping criteria based on the learning curve~\cite{Domhan2015} or MRS~\cite{Camero2018lowcost}) to \emph{evolving} the architecture of the network (neuroevolution).

On the other hand, when it comes to RNN optimization, there are two particular issues: the exploding and the vanishing gradient~\cite{bengio1994learning}. Many alternatives have been proposed to tackle with this problems~\cite{Pascanu2013}. One of the most popular ones is the Long Short Term Memory (LSTM) cell~\cite{hochreiter1997long}. However, in spite of its ability to effectively deal with these issues, the problem still remains open, because the learning process is also affected by the weight initialization strategy~\cite{ramos2017quantitative} and the algorithm parameters~\cite{haykin2009neural}.

\subsection{Neuroevolution}
\label{subsec:neuroevolution}

Neuroevolutionary approaches typically represent the DNN architecture as solution candidates in specifically designed variants of state-of-the-art evolutionary algorithms.
For instance, genetic algorithms (GA) have been applied to evolve increasingly complex neural network topologies and the weights simultaneously, in the so-called NeuroEvolution of Augmenting Topologies (NEAT) method~\cite{larochelle2009exploring,stanley2002evolving}. However, NEAT has some limitations when it comes to evolving RNNs~\cite{miikkulainen2019evolving}, e.g., the fitness landscape is deceptive and a large number of parameters have to be optimized. For RNNs, NEAT-LSTM~\cite{rawal2016evolving} and CoDeepNeat~\cite{Liang:2018:EAS:3205455.3205489} extend NEAT to mitigate its limitations when evolving the topology and weights of the network. Besides NEAT, there are several evolutionary-based approaches to evolve an RNN, such as EXALT~\cite{elsaid2019evolving}, EXAMM~\cite{ororbia2019investigating}, or a method using ant colony optimization (ACO) to improve LSTM RNNs by refining their cellular structures~\cite{ElSaid2018}. 

A recent work~\cite{camero2019specialized} suggested to address the issue of huge training costs when evolving the architecture. In that research, the objective function, that is usually evaluated by training the candidate network on the full data set evolved by a complete training of the candidate network, instead it is approximated by the so-called MAE random sampling (MRS) method, in which no actual training is required. In this manner, the time required for a function evaluation is drastically reduced in the architecture optimization process.

\subsection{Mean Absolute Error Random Sampling}
\label{subsec:MRS}

MAE Random Sampling is an approach to evaluate the expected error performance of a given architecture. First, the weights of the network are randomly initialized. Second, the error is calculated (i.e., the real and expected output are compared). This two-step process is repeated, and the errors are accumulated. Then, a probabilistic density function (e.g., a truncated normal distribution) is fitted to the error values. Finally, the probability of finding a set of weights whose error is below a user-defined \emph{threshold} is estimated. In other words, by using a random sampling of the output (error), we are estimating how \emph{easy} (i.e., a high probability) it would be to find a \emph{good} (i.e., small error) set of weights.

Given a training data set $\mathcal{D} = \{(\mathbf{x}_i, y_i)\}_i^N, \mathbf{x}_i\in\mathbb{R}^n$, for a given network architecture $\mathbf{h}\in\mathcal{H}$ and $Q$ i.i.d. random weight matrices $\{\mathbf{W}_i\}_{i=1}^Q, \mathbf{W}_i \sim \bm{\mathcal{N}}(\mathbf{0}, \mathbf{I})$, the \emph{Mean Absolute Error} (MAE) of this RNN is denoted as $\mathcal{E} = \{\operatorname{MAE}(\mathcal{D}, \mathbf{h}, t, \mathbf{W}_{i})\}_{i=1}^Q$, where $t$ is the number of time steps in the past used for the prediction. Let $\mu$ and $\sigma$ denote the sample mean and standard deviation of the error sample $\mathcal{E}$. Then the so-called \emph{Mean Absolute Error Random Sampling} (MRS) measure is defined as the empirical probability of obtaining a better error rate than a user-specified threshold $p_{\text{m}}$:
\begin{equation} \label{eq:MRS}
    \textsc{mrs}(\mathcal{D}, \mathbf{h}, t, p_{\text{m}}, Q) = \frac{\Phi\left(\frac{p_m-\mu}{\sigma}\right) -\Phi\left(-\frac{\mu}{\sigma}\right)}{1- \Phi\left(-\frac{\mu}{\sigma}\right)},
\end{equation}
where $\Phi$ stands for the cumulative distribution function (CDF) of the standard normal distribution. The MRS value is calculated from a truncated normal distribution (on the interval $[0,\infty)$), whose location and scale parameters are set to the sample mean $\mu$ and standard deviation $\sigma$, respectively. Throughout this paper, we shall set $p_{\text{m}}$ to $1\%$. Intuitively, the higher MRS value a network architecture that yields a higher MRS value would be more likely to possess a much smaller (hence better) MAE rate after the backpropagation training. Hence, it seems promising to use MRS as a training-free estimation for the performance of neural networks. 

In this paper, we shall adopt MRS as the objective function (that is subject to maximization) for the architecture optimization. For a detailed discussion of MRS, please refer to~\cite{Camero2018lowcost}.

\subsection{Bayesian Optimization} 
\label{subsec:bayesopt}

The so-called \emph{Bayesian Optimization} (BO) (a.k.a.~Efficient Global Optimization)~\cite{jones1998efficient,movckus1975bayesian} algorithm has been applied extensively for automated algorithm configuration tasks~\cite{1554761,horn2015model,Hutter2011}. Bayesian optimization is a sequential global optimization strategy that does not require the derivatives of the objective function and is designed to tackle expensive global optimization problems. Given a real-valued \emph{maximization problem} $f\colon \mathcal{H}\rightarrow \mathbb{R}$ (e.g., $f=\textsc{mrs}$ in the following), BO employs a surrogate model, e.g., Gaussian process regression (GPR) or random forests (RF), to approximate the landscape of the objective function, which is trained on an initial data set $(X, Y)$. Here, $X\subset\mathcal{H}$ is typically sampled in the search space $\mathcal{H}$ using the Latin Hypercube Sampling (LHS) method and $Y=\{f(\mathbf{h})\colon \mathbf{h}\in X\}$ is the set of function values of points in $X$. Essentially, the prediction from surrogate models and the estimated prediction uncertainty are considered simultaneously to propose new candidate solutions for the evaluation. Loosely speaking, the model prediction and its uncertainty are taken as input to the so-called \emph{acquisition function} (or \emph{infill criterion}), which can be interpreted as the utility of unseen solutions and hence is subject to maximization when proposing new candidate solutions. An example of commonly used acquisition functions is the Expected Improvement (EI)~\cite{movckus1975bayesian}. Given the predictor $m\colon \mathcal{H}\rightarrow \mathbb{R}$, the uncertainty of predictions $s(\mathbf{h})\coloneqq \mathbb{E} \{(m(\mathbf{h}) - f(\mathbf{h}))^2\}$ of the surrogate model and the current best function value $y_{\text{max}}=\max\{Y\}$, the EI criterion can be expressed for an unknown point $\mathbf{h}\in\mathcal{H}$:
\begin{align}\label{eq:EI}
\scriptsize
	\operatorname{EI}(\mathbf{h}) & =I(\mathbf{h})\Phi\left(\frac{I(\mathbf{h})}{s(\mathbf{h})}\right) + s(\mathbf{h})\phi\left(\frac{I(\mathbf{h})}{s(\mathbf{h})}\right),
\end{align}
where $I(\mathbf{h}) = m(\mathbf{h}) - y_{\max}$ and where $\phi$ stands for the probability density function (PDF) of the standard normal distribution. Note that the new candidate solution is generated by maximizing the EI criterion, namely
\begin{equation}\label{eq:EI-maximization}
	\mathbf{h}^* = \argmax_{\mathbf{h}\in\mathcal{H}}\operatorname{EI}(\mathbf{h}).
\end{equation}

After evaluating the new candidate solution $\mathbf{h}^*$, $\mathbf{h}^*$ and its objective function value are included in the data set $(X,Y)$ and the surrogate model will be re-trained. Please, see~\cite{8122656} for an overview of the acquisition functions.

Despite being a proven technique for automated algorithm configuration tasks~\cite{1554761,horn2015model,Hutter2011}, the state-of-the-art of BO does not reconcile well with variable-length solution problems~\cite{kim2019bayesian,shahriari2015taking}. Therefore, in this study we propose multiple strategies to cope with variable-length solutions (inherent to the architecture search problem).

\subsection{Our contribution}
Herein, we briefly summarize the novelty of the architecture search described in the following sections and compare those to the state-of-the-art works reviewed in this section.
\begin{itemize}
    \item We propose to use the Mean Absolute Error Random Sampling (MRS) procedure as the objective function for the architecture search, which is relatively much inexpensive compared to full training of the same architecture on the same data. In contrast to employing full training, e.g.,~\cite{rawal2016evolving}, our approach could allow for more iterations of the Bayesian optimization algorithm.
    \item We designed three different encoding schemes that turn the neural architecture search that is inherently a variable-dimensional problem(for instance, the NEAT~\cite{larochelle2009exploring} approach operates on the network topology directly) into an optimization problem with fixed dimensions, hence facilitating the application of surrogate modeling and Bayesian optimization accordingly. 
    \item We contemplated making the Bayesian optimization more efficient and effective by imposing a penalty on the infeasible solutions or warm-starting the search process with infeasible solutions.
\end{itemize}

\section{The proposed approach}
\label{sec:approach}

In this paper, it is proposed to optimize the architecture of an RNN by a combination of Bayesian optimization (BO) and Mean Absolute Error Random Sample (MRS) to reduce the running time of the architecture search. Specifically, this is to solve the following problem using \emph{Bayesian optimization},
\begin{equation}
\argmax_{\mathbf{h}\in\mathcal{H}}\textsc{mrs}(\mathcal{D}, \mathbf{h}, p_{\text{m}}, Q),
\end{equation}
given a training data set $\mathcal{D}$, a cutoff threshold $p_{\text{m}}$ and the number of random weights used in MRS. Importantly, as the architecture could shrink and expand in the search, its natural representation takes a variable-length form, which does not reconcile well with the state-of-the-art BO algorithm. To resolve this issue, three fixed-length encoding schemes are proposed to represent network architectures with variable sizes. Note that in this paper the search space $\mathcal{H}$ is determined by each encoding scheme (please see below). Also, we only employ the random forest model in the Bayesian optimization procedure (described in Alg.~\ref{alg:bo}) for the following reason: the design space of neural architecture comprises of integer/Boolean variables, which can be dealt with naturally by random forests. Gaussian process regression, which works over Euclidean spaces, is by default not applicable in this scenario. Although there are many recently endeavours in extending GPR's ability to handle the discrete and integer variables (e.g.,~\cite{Nguyen0RSV20}), it is not our major aim herein to compare the performance of Bayesian optimization when coupled with different surrogate models and hence we decided to choose the simplest random forest model to validate the proposed algorithm.

\subsection{Encoding Schemes}
\label{subsec:encoding}

Assuming that the number of neurons per each layer is restricted to the range $[\munderbar{N}..\bar{N}]$, the number of layers is $m\in[\munderbar{M}..\bar{M}]$, and $T$ denotes the maximum number of steps taken in back-propagation throughout time, three encoding schemes are proposed in this paper:
\begin{itemize}
    \item \textbf{Plain}: the total length of this encoding is $m +1$.
    $$\mathbf{h}=\left[h_1,  h_2, \ldots, h_m, l\right] \in \left(\{0\} \cup [\munderbar{N}..\bar{N}]\right)^m \times [1..T],$$
    where $h_i$ is the number of neurons per each layer and $l$ is the number of time steps. Note that $h_i$ can take value zero, meaning there is no neuron in this layer and hence it is effectively dropped in the decoding procedure.
    \item \textbf{Flag}: the total length of this encoding is $2m + 1$.
    $$\mathbf{h}=\left[h_1, b_1, h_2, b_2, \ldots, h_m, b_m, l\right] \in [\munderbar{N}..\bar{N}]^m \times \{0, 1\}^m \times [1..T],$$
    where $b_i \in \{0, 1\}$ is the so-called ``flag'' that disables layer $h_i$ if $b_i = 0$ when decoding such a representation to compute the actual architecture.
    \item \textbf{Size}: the total length of this encoding is $m + 2$.
    $$\mathbf{h}=\left[h_1,  h_2, \ldots, h_m, s, l\right] \in [\munderbar{N}..\bar{N}]^m \times [1..m] \times [1..T],$$
    where $s\leq m$ is the number of layers from the start of the representation that are considered in decoding, namely only $h_1, h_2, \ldots, h_s$ are used to generate the actual architecture. 
\end{itemize}

\begin{figure}[!htbp]
 \centering
 \begin{tikzpicture}[scale=.8, every node/.style={scale=.8}]
     \matrix[nodes={draw, fill=blue!20, minimum size=10mm, line width=1pt},
         row sep=1mm,column sep=0cm] {
     \node[draw=none,fill=none] {Plain};&
     \node[square] {$h_1$}; &
     \node[square] {$h_2$}; &
     \node[square] {$\ldots$}; &
     \node[square] {$h_m$}; &
     \node[square] {$l$}; \\
     \node[draw=none,fill=none] {Size};&
     \node[square] {$h_1$}; &
     \node[square] {$h_2$}; &
     \node[square] {$\ldots$}; &
     \node[square] {$h_m$}; &
     \node[square] {$s$}; &
     \node[square] {$l$}; \\
     \node[draw=none,fill=none] {Flag};&
     \node[square] {$h_1$}; &
     \node[square] {$b_1$}; &
     \node[square] {$\ldots$}; &
     \node[square] {$h_m$}; &
     \node[square] {$b_m$}; &
     \node[square] {$l$}; \\
     };
 \end{tikzpicture}
 \caption{Illustrations of the proposed encoding schemes.}
 \label{figure:encodings}
\end{figure}
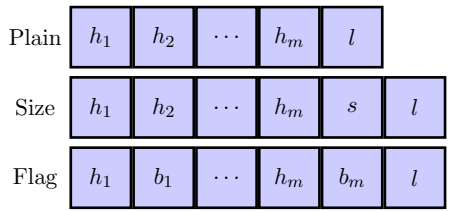

We shall use the notation $\texttt{code} \in \{\text{plain},\text{flag},\text{size}\}$ for the encoding scheme henceforth. In this manner, a fixed-length representation can be used to optimize variable-size architectures. For each case, in the decoding procedure, an output layer is appended to the RNN structure encoded in the search algorithm, to match the expected output dimension. Note that the activation function of the output layer has to be set according to the type of the task in each problem.

\subsection{Decoding}
\label{subsec:valid}

It is worthwhile to note that the decoding procedure of all three representations is a \emph{many-to-one mapping}. For instance, given a plain representation with a maximum of five layers ($m = 5$), $[h_1,h_2,0,0,h_5,l]$ and $[h_1,h_2,h_5,0,0,l]$ are representing exactly the same architecture. If $[h_1,h_2,h_5,0,0,l]$ has already been evaluated in the optimization process, then assessing the performance of $[h_1,h_2,0,0,h_5,l]$ is purely redundant. To determine the equivalence among representations, it is necessary to apply appropriate decoding functions for each type of representation: 
\begin{equation} \label{eq:decoding}
    \texttt{decode}(\mathbf{h}) = 
    \begin{cases}
        \text{keep } h_i \text{ if } h_i > 0, i=1,\ldots,m. &\text{if the plain encoding} \\
        \text{keep } h_i \text{ if } b_i =1, i=1,\ldots,m. &\text{if the flag encoding}\\
        \mathbf{h} \mapsto [h_1, h_2, \ldots, h_s, l] &\text{if the size encoding}
    \end{cases}
\end{equation}
As the decoding function is a many-to-one mapping, the BO algorithm could potentially propose the same architecture constantly (even with different representations before decoding), and hence the search efficiency would be drastically affected due to the following facts 1) the convergence of BO would be hampered as such an iteration (where the seen architecture is proposed again) makes no progress and the there is no information gain for the surrogate model therein, and 2) the same network architecture has to be evaluated again by MRS, which is wasteful even if MRS is much more efficient as compared the full network training.
To cope with the former issue, it is important to avoid proposing the same architecture again as much as possible. In this study, we propose two alternative strategies which both rely on the definition of \emph{``infeasibility''} (please see below) for representations:
\begin{itemize}
	\item to set the MRS value of infeasible representations to the worst possible value (zero), which will be learned by the surrogate model underlying BO. Hence, the infeasible ones would not likely to be proposed by the surrogate model, or
	\item to use the original MRS values (as in Eq.~\eqref{eq:MRS}) and add constraints on the EI criterion to screen out infeasible representations. Note that in this case the surrogate model will be built on the original MRS values.
\end{itemize}
For the latter, the simplest solution is to maintain a lookup table to register the architectures (together with objective values) that are evaluated before.

\paragraph{Infeasible representation}
Taking the plain encoding scheme as an example, a representation taking the form $[h_1, \ldots h_q, 0, \ldots,0, l]$ (where $h_i >0$) shall be called \emph{feasible}, e.g., $[h_1,h_2,h_5,0,0,l]$ is an infeasible representation when $m=5$. $[h_1,h_2,h_5,0,0,l]$ represents the same architecture with the other $16$ representations (by inserting two zeros at four different positions, e.g., $[h_1,h_2,0,0,h_5,l]$ and $[h_1,0,h_2,0,h_5,l]$). The other representations shall be called \emph{``infeasible''}, which will be assigned with a fixed objective value that is worse than all the feasible solutions. Particularly, since we are maximizing MRS (which is a probability value), we set the penalized objective function value to be equal to zero. The rationale behind this treatment is that whenever the Bayesian optimization (BO) algorithm proposes an infeasible representation, the penalized objective function value will be learned by the surrogate model of BO and hence the chance of generating such representations will diminish gradually. In this manner, we are guiding the optimization process through the feasible ones and thus the search space is virtually reduced. Note that the BO algorithm still needs to make lots of infeasible trials before it stops proposing the infeasible ones, due to the large combinatorial space. It is conceptually better to directly avoid generating such representations by a constraint handling method (see below). The idea of defining the infeasible representation can be easily extended to the flag encoding scheme by \emph{masking} $h_i$ with $b_i$, i.e., replacing the value of $h_i$ with a zero if $b_i$ is equal to zero. However, this idea can not be applied to the \emph{size} encoding scheme. 

\paragraph{Constraint handling} To avoid generating infeasible representations, we propose to assign penalty values to infeasible ones and to use a constraint handling method when proposing new candidate representations in BO. 
In addition, representations that are already evaluated will be also be penalized by the length of itself (the maximum penalty at line $4$). For an infeasible representation that has \emph{not} been evaluated (line $5$), the number of zeros located before the last nonzero element is used as the penalty value. In line $7$, the decoded representation is registered in a set $L$ to check whether a representation has been evaluated before. The penalty value will be added to the EI criterion when proposing the candidate representations (see line $13$ of Alg.~\ref{alg:bo}). As for the constraint handling, a \emph{dynamic penalty} method is adopted here, where the penalty value will be scaled up with increasing iterations of BO. We choose the dynamic penalty here because it yields a relatively small penalty in the early phase of the search, allowing for exploring the infeasible regions within the search space, which is particularly critical to move between disconnected feasible regions. Also, as the search iteration increases, the penalty value will be enlarged to ensure a feasible solution as the outcome. In this manner, the following penalized infill criterion is used to propose candidate representations (instead of Eq.~\ref{eq:EI-maximization}):
\begin{equation} \label{eq:constrained-EI}
    \mathbf{h}^* = \argmax_{\mathbf{h}\in \mathcal{H}}\operatorname{EI}(\mathbf{h} ; \mathcal{M}) - Ct\cdot\textsc{penalty}(\mathbf{h}, X),
\end{equation}
where 1) $X$ is a set containing all evaluated solutions (not decoded), 2) $t$ is the iteration counter of BO, and 3) $C=0.5$ is a scaling factor. The intuition of this treatment is that the penalty value would have a large impact on the maximization of EI in the late stage, such that the probability of generating infeasible solutions becomes marginal. Also, the penalty value of $\mathbf{h}$ equals its length when it has been evaluated before, i.e., $\mathbf{h}\in X$, for avoiding proposing duplicated solutions, and otherwise, it is set to penalize $\mathbf{h}$ by the number of zeros preceding non-zero elements thereof, namely,
\begin{equation} \label{eq:penalty}\scriptsize
\textsc{penalty}(\mathbf{h}, X)=
\begin{cases}
  \texttt{length}(\mathbf{h}), &\text{ if }  \mathbf{h} \in X \\ 
  |\left\{h_i \colon \forall i \in [1..n-1]\left(h_i = 0 \cap\exists j \in [i+1..n](h_j = 1)\right)\right\}|, &\text{ otherwise}.
\end{cases}
\end{equation}
		

\subsection{A Warm-start Strategy}
\label{subsec:warm-start}

Within the Bayesian optimization algorithm, a surrogate model (e.g., a random forest) is used to learn the mapping from the evaluated solutions to the corresponding objective values. Typically, the Bayesian optimization starts with initializing the surrogate model by some randomly generated solutions. The basic idea of the so-called ``warm-start'' strategy is to augment the initial solutions by a set of infeasible solutions that can be generated before the optimization, such that the optimization process is started with a priori information. The infeasible solutions can be generated by randomly picking some components of a solution and setting them to zero for both the plain and flag encoding. Additionally, the objective value of those infeasible solutions is assigned with some default bad value (it is set to zero here since the MRS measure, which is the objective function of the architecture search, is bounded by zero from below), without the need to execute the MRS procedure. We anticipate that this warm-start strategy will add a bias in proposing the new candidate solutions in BO, steering the optimization process away from the infeasible solutions. 

In all, the pseudo-code of the proposed approach is described in Alg.~\ref{alg:bo}. After creating the initial data set of BO $(X,Y)$ using Latin Hypercube Sampling~\cite{mckay1979comparison}, the user can choose to turn on the generation of the warm-data prior to the optimization loop (lines $6$-$9$). A set $X'$ consisting of decoded representations is meant to track all the unique architectures that have been evaluated in MRS (line $11$). In line $16$, the constrained EI maximization is applied if the constraint method is enabled. The newly proposed solution $\mathbf{h}^*$ is decoded (line $20$), after which we check if its decoded counterpart ${\mathbf{h}^*}'$ has been evaluated (line $21$). If ${\mathbf{h}^*}'$ is not evaluated before (line $22$-$28$), the feasibility of $\mathbf{h}$ is then checked and its objective value is set to zero in case of being infeasible (Otherwise, we evaluate its decoded representation ${\mathbf{h}^*}'$ in MRS (line $26$)) If ${\mathbf{h}^*}'$ has been evaluated before, its objective value is looked up in the data set $(X,Y)$ (line $30$ and $31$). The newly proposed candidate representation and its objective value are appended to BO's data set $(X, Y)$ (lines $33$ and $34$). Afterwards, the random forest model is re-trained on the augmented data set (line $35$).

\begin{algorithm}[!htp]
	\caption{Efficient Architecture Optimization for RNNs}\label{alg:bo}
	\begin{algorithmic}[1]
		\State \textbf{input}: A data set $\mathcal{D}$, an encoding scheme $\texttt{code} \in \{\text{plain},\text{flag},\text{size}\}$, the random forests algorithm \textsc{rf}, and the maximal iteration number
		$t_{\text{max}}$.
		\State \textbf{output}: a full training RNN model
		\State{$C \leftarrow 0.5, t \leftarrow 0, p_{\text{m}} \leftarrow 0.01, Q\leftarrow 100$}
		\State{Determine the search space $\mathcal{H}$ according to \texttt{code}}
		\State{Generate $X\subseteq \mathcal{H}$ using Latin Hypercube Sampling} 
		\State{$Y \leftarrow \{\textsc{mrs}(\mathcal{D}, \texttt{decode}(\mathbf{h}), t, p_{\text{m}}, Q)\colon \mathbf{h}\in X\}$}\Comment{evaluate $X$}
		\If{``warm-start'' is enabled}
			\State{generate the warm data $(X_{\text{warm}},Y_{\text{warm}})$} \Comment{See sec.~\ref{subsec:warm-start}}
			\State{$X\leftarrow X\cup X_{\text{warm}}, Y\leftarrow Y\cup Y_{\text{warm}}$}
		\EndIf
		\State{$X' \leftarrow \{\texttt{decode}(\mathbf{h})\colon \mathbf{h} \in X\}$} \Comment{set of evaluated architectures}
		\State{$\mathcal{M}\leftarrow\textsc{rf}(X, Y)$}\Comment{surrogate model training}
		\While{$t < t_{\text{max}}$}
			\If{``constraint-handling'' is enabled}
				\State{$\mathbf{h}^* \leftarrow \argmax_{\mathbf{h}\in \mathcal{H}}\operatorname{EI}(\mathbf{h} ; \mathcal{M}) - Ct\cdot\textsc{penalty}(\mathbf{h}, X)$} \Comment{penalized EI}
			\Else
	 			\State{$\mathbf{h}^* \leftarrow \argmax_{\mathbf{h}\in \mathcal{H}}\operatorname{EI}(\mathbf{h} ; \mathcal{M})$} \Comment{unconstrained case}
	 		\EndIf
	 		\State{${\mathbf{h}^*}' \leftarrow \texttt{decode}(\mathbf{h}^*)$}\Comment{solution decoding (Eq.~\eqref{eq:decoding})}
	 		\If{${\mathbf{h}^*}' \notin X'$} \Comment{for unseen architectures}
	 		    \If{``infeasible-solution'' is enabled \textbf{and} \\ \hspace{1.4cm}$\texttt{code} \neq \text{size}$ \textbf{and} $\mathbf{h}^*$ is infeasible}
	 		        \State{$y^* \leftarrow -\textsc{inf}$}  \Comment{penalty value for the infeasible ones}
	 		    \Else
	 		        \State{$y^* \leftarrow \textsc{mrs}(\mathcal{D}, {\mathbf{h}^*}', t, p_{\text{m}}, Q)$} \Comment{evaluate ${\mathbf{h}^*}'$ using $\textsc{mrs}$}
	 		    \EndIf
	 		    \State{$X'\leftarrow X'\cup \{{\mathbf{h}^*}'\}$} \Comment{add to the set of evaluated architectures}
	 		\Else
	 		    \State{$S \leftarrow \left\{y \colon \forall(\mathbf{h}, y) \in (X, Y) \land \texttt{decode}(\mathbf{h}) = {\mathbf{h}^*}'\right\}$} \Comment{the objective value of evaluated solutions that decodes to the same architecture as ${\mathbf{h}^*}'$}
	 		    \State{$y^* \leftarrow$ sample a value from $S$ uniform at random}
	 		\EndIf
	 		\State{$X\leftarrow X\cup \{\mathbf{h}^*\}, Y \leftarrow Y \cup \{y^*\}$}\Comment{augment the data set}
	 		\State{$\mathcal{M}\leftarrow\textsc{rf}(X, Y)$} \Comment{re-train the random forest model}
	 		\State{$t\leftarrow t + 1$}
	 	\EndWhile
	 	\State $y_{\text{best}} \leftarrow \max\{Y\}$ and $\mathbf{h}_{\text{best}}$ is the corresponding solution to $y_{\text{best}}$
	 	\State $\mathbf{h}_{\text{trained}} \leftarrow \operatorname{ADAM}(\mathcal{D}, \mathbf{h}_{\text{best}})$ \Comment{train the final neural architecture}
	 	\State \textbf{return} $\mathbf{h}_{\text{trained}}$
	\end{algorithmic}
\end{algorithm}

\section{Experiments}
\label{sec:experiments}

In this section, we present the experimental study performed to test the proposed approach. First, we present the three prediction problems used to benchmark the method. Second, we present the experimental setup and the results of several combinations of the three strategies presented, i.e., infeasible solution, warm start, constraint handling, and encoding. Later, we compare the time between MRS and (short training) Adam. Finally, we study the error trade-off while changing the number of MRS samples.

\subsection{Data sets}
\label{subsec:dataset}

We tested the approach on three prediction problems: \emph{sine wave}, \emph{waste}~\cite{Ferrer2018}, and \emph{load forecast}~\cite{chen2004load}.

\paragraph{The \emph{sine wave}} is a mathematical curve that represents a periodic oscillation. Despite its simplicity, it is extensively used to analyze systems~\cite{bracewell1986fourier}. It is usually expressed as a function of time (\emph{t}), where $A$ is the peak amplitude, $f$ the frequency, and $\phi$ the phase (Equation~\ref{equation:sine}). Its study is interesting because, by adding sine waves, it is possible to approximate any periodic waveform~\cite{bracewell1986fourier}. We sampled the sine wave described by: $A=1$, $f=1$, and $\phi=0$, in the range $t \in [0,100]$~seconds, and at 10 samples per second. Then, given a truncated part of the time series (i.e., a time steps number of points of the sampled sine wave), the problem consists in predicting the next value. 
\begin{align}\label{equation:sine}
y(t) = A \sin(2 \pi f  t + \phi)
\end{align}

\paragraph{The \emph{waste} problem} introduced in~\cite{Ferrer2018}, consists of predicting the filling level of 217 recycling bins located in the metropolitan area of a city in Spain, recorded daily for one year. Thus, given the historical filling levels of all containers (217 input values per day), the task is to predict the next day (i.e., the filling level of all bins). It is important to notice that this problem has been used as a benchmark in several studies~\cite{ferrer2019bin,camero2019waste,camero2019specialized} and that it is a real-world problem.

\paragraph{The \emph{load forecast} problem} provided by the European Network on Intelligent Technologies for Smart Adaptive Systems (EUNITE, \url{http://www.eunite.org}) as part of a competition~\cite{chen2004load,lang2018short}, is a data set consisting of the electricity load demand of the Eastern Slovakian Electricity Corporation. It was recorded every half hour, from January 1, 1997, to January 31, 1999. Also, the temperature (daily mean) and the working calendar for this period are provided. Then, based on this data, the challenge is to predict the next maximum daily load. In other words, given the load demand (52 variables), i.e., the load demand recorded every half hour (48), the max daily load (1), the daily average temperature (1), the weekday (1), and the working day information (1), the task is to predict the max daily load of the next day (1). Note that the last month is used as the test data, thus our results may be compared directly against the competitors.

\subsection{Performance}
\label{subsec:performance}

We implemented our approach\footnote{Code available in \url{https://github.com/acamero/dlopt}} in Python 3, using DLOPT~\cite{camero2018dlopt}, MIP-EGO~\cite{wang2018cooling}, Keras~\cite{chollet2015keras}, and Tensorflow~\cite{abadi2016tensorflow}. We used LSTM cells to build the decoded stacked architectures (as a way to mitigate the exploding and vanishing gradient problems~\cite{hochreiter1997long}), and Adam truncated through time~\cite{werbos1990backpropagation} (i.e., sharing all parameters in the unfolded models) to train the final solutions, with default parameter values~\cite{kingma2014adam}.

We defined the search space (i.e., the constraints to the RNN architectures) of the three problems studied (Table~\ref{table:search-spaces}) according to the datasets and the state-of-the-art. Particularly, the sine wave search space is taken from~\cite{camero2019specialized} and the waste search space is copied from~\cite{camero2019waste} to enable a direct comparison.

\begin{table}
    \caption{Optimization search spaces}
	\centering 
    \begin{tabular}{ lrrr }
    \hline
    Parameter & Load Range & Sine Range & Waste Range \\
    \hline
    Hidden layers (M)     & [1,8]     & [1,3]   & [1,8] \\
    Look back (T)         & [2,30]    & [2,30]  & [2,30] \\
    Neurons per layer (N) & [10,100]  & [1,100] & [1,300] \\
    \hline
	\end{tabular} \\    
    \label{table:search-spaces}
\end{table}


Also, to ease the visualisation of the results, we defined the following naming scheme to denote different combinations of encoding, warm start, invalid, and the constraint handling method: \\ $\small \texttt{[constraint][warm start][infeasible][encoding]}$.

Specifically, we use a character to denote each variant: Constraint (\texttt{C}), Warm start (\texttt{w}), Infeasible (\texttt{I}), and Encoding (\texttt{F}: flag, \texttt{S}: size, and \texttt{P}: plain). A dash (\texttt{-}) means that the corresponding alternative was not used. For example, \texttt{-W-F} corresponds to the combination of warm data and the flag encoding (i.e., without constraint handling and without invalid solution penalty).

Finally, we execute $30$ independent runs for each combination of encoding, warm start, and the constraint handling method on a heterogeneous Linux cluster with more than 200 cores and 700 GB RAM, managed by HTCondor. In these experiments we used the optimization parameter values presented in Table~\ref{table:optimization-params}. The remainder of this subsection introduces the performance results for the three problems and some insights into the solutions.

\begin{table}[h]
    \caption{BO and MRS parameter values}
	\centering 
    \begin{tabular}{ lrclr }
    \cline{1-2} \cline{4-5}
    Parameter & Value &
        & Parameter & Value \\
    \cline{1-2} \cline{4-5}
    No. Samples (Q) & 100 &
        & Threshold ($p_m$)   & 0.01 \\
    Max Evaluations   & 100 &
        & Init Solutions & 10 \\
    Epochs  & 1000  &
        & Dropout   & 0.5 \\
    \cline{1-2} \cline{4-5}
	\end{tabular} \\    
    \label{table:optimization-params}
\end{table}

Note that the parameters presented in Table~\ref{table:search-spaces} and~\ref{table:optimization-params} were taken from~\cite{camero2019specialized,camero2019waste}. We decided to chose these values (instead of performing an hyperparameter tuning) to enable a direct comparison with our competitors.

\subsubsection{Sine Wave}
The range of the sine function is $[0,1]$, thus we set the activation function of the dense output layer to be a \texttt{tanh}. Due to the immense number of \emph{invalid} solutions, we implemented a \emph{limited} version of the infeasible solution listing, i.e., instead of enumerating all infeasible solutions, we list a subset of them. Particularly, we listed the infeasible solutions described by the min and max values of each parameter (i.e., the number of neurons per layer and look back). Thus, we added 80 infeasible solutions to the warm-start.

\begin{sidewaystable}
	\centering 
    \caption{Sine optimization results (MAE of the best solution). Groups sharing a letter in the Conover row are not significantly different}        \label{table:sin-summary}
    \begin{tabular}{ lrrrrrrrrrrrr }
    \hline
    	& GDET	& MLES	& ---F	& --IF	& -W-F	& -WIF	& C--F	& CWIF	& ---S	& C—S	& ---P	& --IP \\
    \hline
    Mean	& 0.1419	& 0.1047	& 0.0785	& 0.0882	& 0.0816	& 0.1119	& 0.0839	& 0.1452	& 0.0857	& 0.0745	& 0.1198	& 0.1363 \\
    Median	& 0.1489	& 0.0996	& 0.0738	& 0.0882	& 0.0772	& 0.0861	& 0.0789	& 0.0935	& 0.0748	& 0.0721	& 0.1170	& 0.1244 \\
    Max	& 0.2695	& 0.2466	& 0.1172	& 0.1266	& 0.1185	& 0.3677	& 0.1276	& 0.5723	& 0.1794	& 0.0962	& 0.1700	& 0.3290 \\
    Min	& 0.0540	& 0.0631	& 0.0449	& 0.0505	& 0.0518	& 0.0492	& 0.0631	& 0.0577	& 0.0584	& 0.0525	& 0.0922	& 0.0665 \\
    Sd	& 0.0513	& 0.0350	& 0.0194	& 0.0182	& 0.0161	& 0.0695	& 0.0154	& 0.1367	& 0.0274	& 0.0109	& 0.0177	& 0.0558 \\

    \hline
    Conover & a & bc & \textbf{d} & ef & \textbf{d}e & bf & \textbf{d}ef & c & \textbf{d}ef & \textbf{d} & a & a \\
    \hline
	\end{tabular} \\    

\bigskip
    \caption{Waste optimization results (MAE of the final solution). Groups sharing a letter in the Conover row are not significantly different}    \label{table:waste-summary}
    \begin{tabular}{ lrrrrrrrrrrrr }
    \hline
    	& Cities	& MLES	& ---F	& --IF	& -W-F	& -WIF	& C--F	& CWIF	& ---S	& C—S	& ---P	& --IP  \\
    \hline
    Mean	  & 0.0728	& 0.0790	& 0.0722	& 0.0821	& 0.0730	& 0.0812	& 0.0728	& 0.0728	& 0.0732	& 0.0725	& 0.0744	& 0.0735 \\
    Median	  & 0.0731	& 0.0728	& 0.0723	& 0.0735	& 0.0725	& 0.0730	& 0.0725	& 0.0725	& 0.0736	& 0.0723	& 0.0737	& 0.0731 \\
    Max	      & 0.0757	& 0.1377	& 0.0791	& 0.1227	& 0.0806	& 0.1231	& 0.0767	& 0.0767	& 0.0756	& 0.0760	& 0.0920	& 0.0883 \\
    Min	      & 0.0709	& 0.0691	& 0.0695	& 0.0691	& 0.0703	& 0.0698	& 0.0692	& 0.0701	& 0.0691	& 0.0688	& 0.0717	& 0.0701 \\
    Sd	      & 0.0012	& 0.0172	& 0.0019	& 0.0156	& 0.0020	& 0.0177	& 0.0018	& 0.0014	& 0.0015	& 0.0016	& 0.0041	& 0.0027 \\

    \hline
    Conover & \textbf{a}bc & \textbf{a}bc & \textbf{a} & bcd & \textbf{a}b & d & \textbf{a}b & \textbf{a}b & bcd & \textbf{a}b & cd & bcd \\
    \hline
	\end{tabular} \\    
\bigskip
    \caption{Optimization results (MAPE of the final solution). Groups sharing a letter in the Conover row are not significantly different}     \label{table:eunite-summary}
    \begin{tabular}{ lrrrrrrrrrrrrr }
    \hline
    	& SVM	& RBF	& WK+	& ---F	& --IF	& -W-F	& -WIF	& C--F	& CWIF	& ---S	& C—S	& ---P	& --IP  \\
    \hline
    Mean	& 2.879	& NA	& NA	& 2.726	& 3.148	& 2.595	& 3.066	& 2.158	& 2.844	& 2.321	& 2.235	& 4.823	& 5.287 \\
    Median	& 2.945	& NA	& NA	& 2.466	& 2.933	& 2.368	& 2.814	& 2.099	& 2.846	& 2.125	& 2.050	& 5.040	& 5.213 \\
    Max	& 3.480	& NA	& NA	& 6.271	& 5.207	& 4.594	& 6.031	& 3.364	& 3.901	& 6.271	& 4.605	& 6.999	& 11.004 \\
    Min	& 1.950	& 1.481	& 1.323	& 1.840	& 1.759	& 1.593	& 1.919	& 1.452	& 2.033	& 1.654	& 1.657	& 3.142	& 3.415 \\
    Sd	& 0.004	& NA	& NA	& 0.888	& 1.013	& 0.765	& 1.000	& 0.440	& 0.515	& 0.774	& 0.564	& 0.884	& 1.727 \\
    
    \hline
    Conover & NA & NA & NA & abc & a & bc & a & \textbf{d} & ab & ce & \textbf{d}e & f & f \\
    \hline
	\end{tabular} \\    
    
\end{sidewaystable}

\begin{figure}[ht]
  \centering
  \includegraphics[width=0.8\columnwidth]{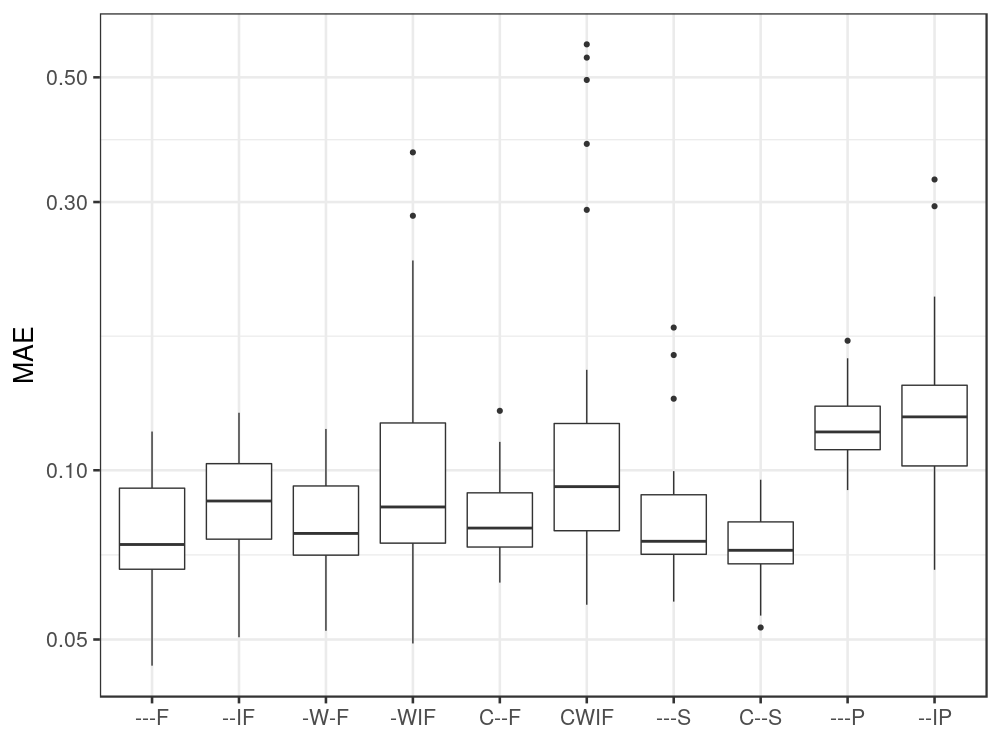}
  \caption{MAE of the sine wave solutions}
  \label{figure:mae-sin}
\end{figure}

Table~\ref{table:sin-summary} summarizes the results of the experiments, where MLES and GDET are the results presented in~\cite{camero2019specialized}, and the other results correspond to the tested combinations. Figure~\ref{figure:mae-sin} shows the distribution of the MAE of the solutions of the sine wave problem. The Friedman rank sum test $p$-value is less than 2.2e-16 (chi-squared = 138.17, df = 11). Therefore, we performed a pairwise comparison using the Conover test for a two-way balanced complete block design~\cite{conover1979multiple}, and the Holm $p$-value adjustment method. The results are presented in the row label Conover in Table~\ref{table:sin-summary}. Groups sharing a letter are not significantly different ($\alpha=0.01$).

The results show that using BO and MRS improves the performance of the final solution (error). On the other hand, multiple combinations of the proposed strategies (i.e., the combinations grouped by the letter \texttt{d}) show a similar performance.


\subsubsection{Waste}
The filling level of the bins ranges from 0 to 1. Accordingly, we set the activation function of the output layer to be a \texttt{sigmoid}. In this case, we added 126976 invalid solutions to the warm start.

\begin{figure}[ht]
  \centering
  \includegraphics[width=0.8\columnwidth]{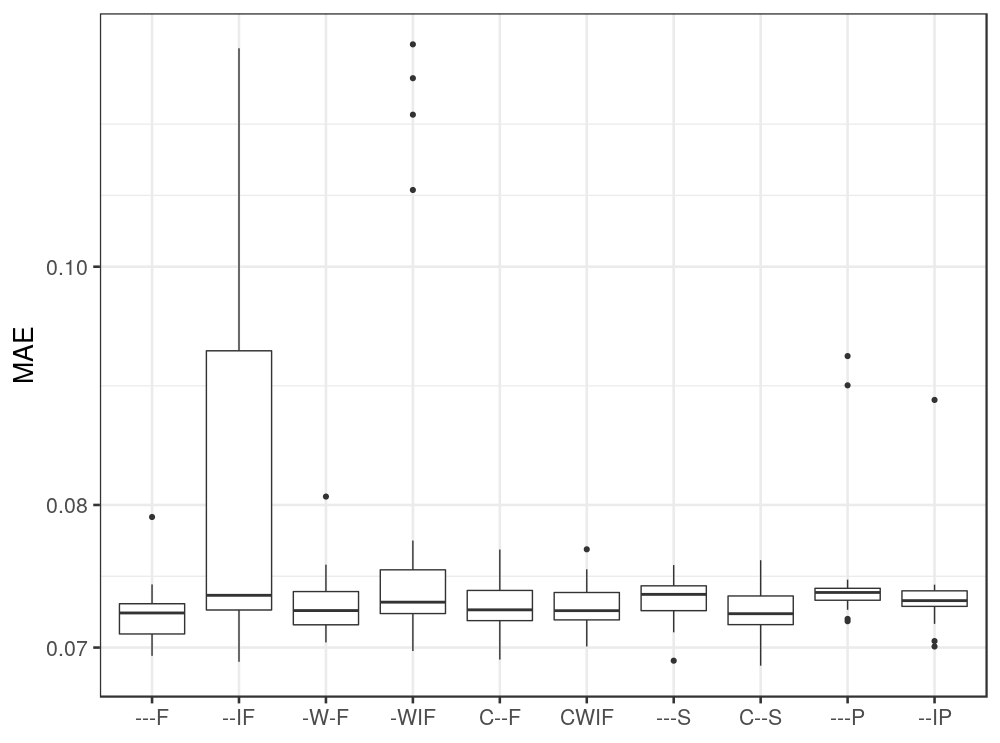}
  \caption{MAE waste}
  \label{figure:mae-waste}
\end{figure}

Table~\ref{table:waste-summary} summarizes the results of the tests on the waste problem. The table also includes the results of \cite{camero2019waste} (Cities) and \cite{camero2019specialized} (MLES). Figure~\ref{figure:mae-waste} shows the distribution of the MAE of the solutions of the waste problem. The Friedman rank sum test $p$-value is equal to 0.02401 (chi-squared = 22.048, df = 11). Therefore, we performed a pairwise comparison using the Conover test for a two-way balanced complete block design~, and the Holm $p$-value adjustment method. The results are presented in the row label Conover in Table~\ref{table:waste-summary}. Groups sharing a letter are not significantly different ($\alpha=0.01$).

In this case, our results are as good as our competitors (the results grouped by the letter \texttt{a}). Nonetheless, it is important to remark that~\cite{camero2019waste} (Cities) trains every candidate solution using Adam, turning out to be more time-consuming.


\subsubsection{Load Forecast}
According to the preprocessing performed in~\cite{lang2018short}, we normalized the data to have a mean equal to zero and a standard deviation equal to one. Then, we set the activation function of the output layer to be \texttt{linear}. Besides, we added 126976 invalid solutions to the warm start.

Table~\ref{table:waste-summary} summarizes our results and the ones presented in 
\cite{chen2004load} (SVM), and \cite{lang2018short} (RBF and WK+, WKNNRW in the original work). In this case, we present the mean absolute percentage error (MAPE) because it is the performance metric used in the referred studies (\texttt{NA} indicates the corresponding data is not available). Figure~\ref{figure:mae-waste} shows the distribution of the MAPE of the solutions of the waste problem. Unfortunately, in this case, we do not have the detailed results of SVM, RBF, and WK+. Thus, we can not perform a detailed analysis considering all competitors. Nonetheless, we performed a detailed analysis considering exclusively the results of our tests. The Friedman rank sum test $p$-value is less than $2.2\times 10^{-16}$ (chi-squared = 146.38, df = 9). Therefore, we performed a pairwise comparison using the Conover test for a two-way balanced complete block design, and the Holm $p$-value adjustment method. The results are presented in the row label Conover in Table~\ref{table:eunite-summary}. Groups sharing a letter are not significantly different ($\alpha=0.01$).


\begin{figure}[ht]
  \centering
  \includegraphics[width=0.8\columnwidth]{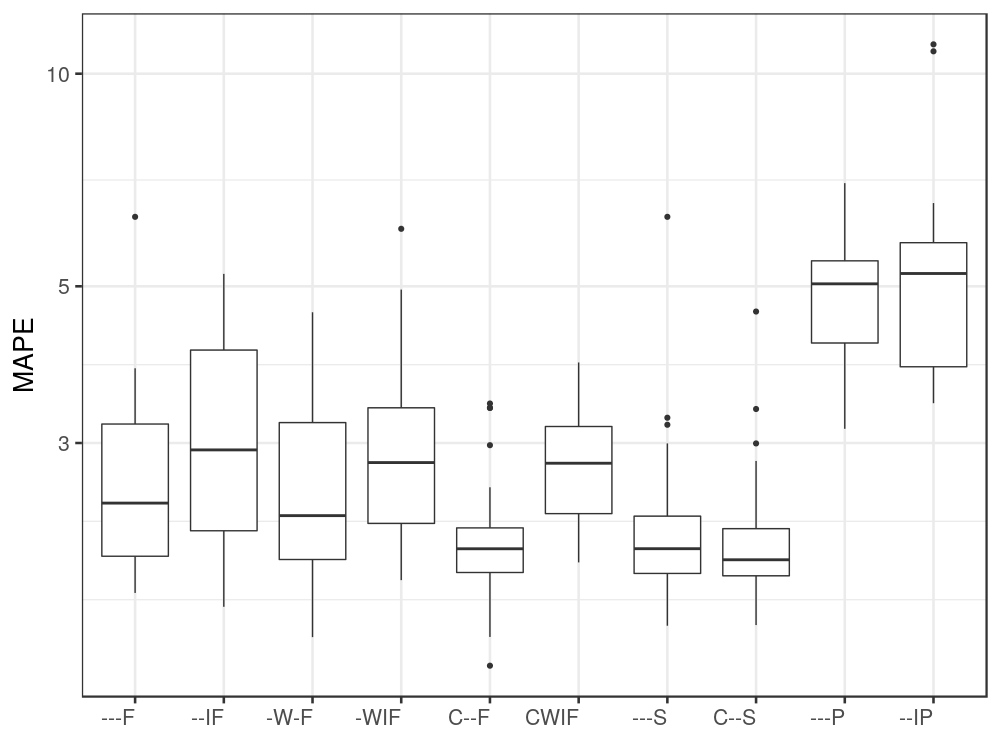}
  \caption{MAPE load forecast}
  \label{figure:mape-eunite}
\end{figure}

\subsubsection{Solutions Overview}

\begin{figure}[ht]
  \centering
  \includegraphics[width=0.8\columnwidth]{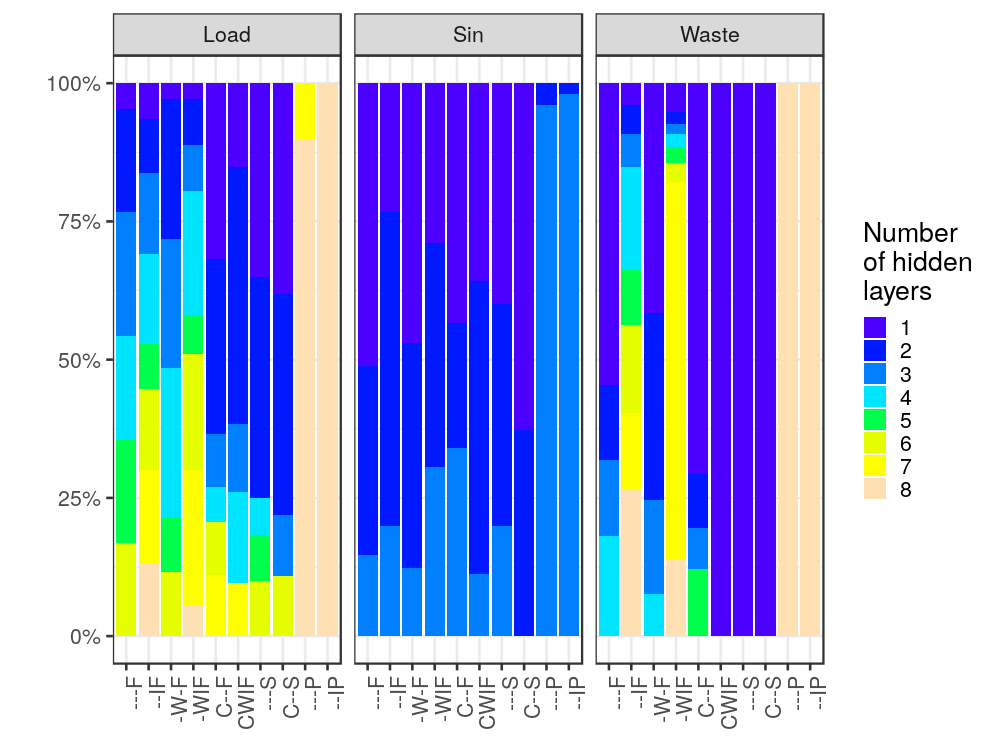}
  \caption{Number of hidden layers of the solutions}
  \label{figure:arch-hl}
\end{figure}

\begin{figure}[ht]
  \centering
  \includegraphics[width=0.8\columnwidth]{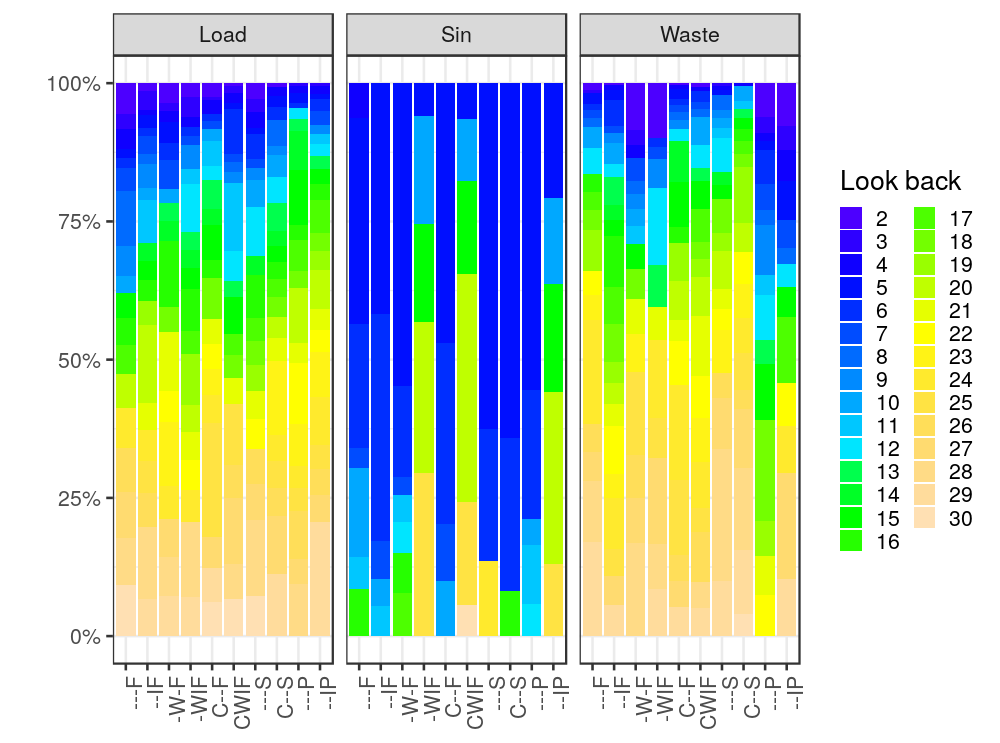}
  \caption{Look back or time steps of the solutions}
  \label{figure:arch-lb}
\end{figure}

\begin{figure}[ht]
  \centering
  \includegraphics[width=0.8\columnwidth]{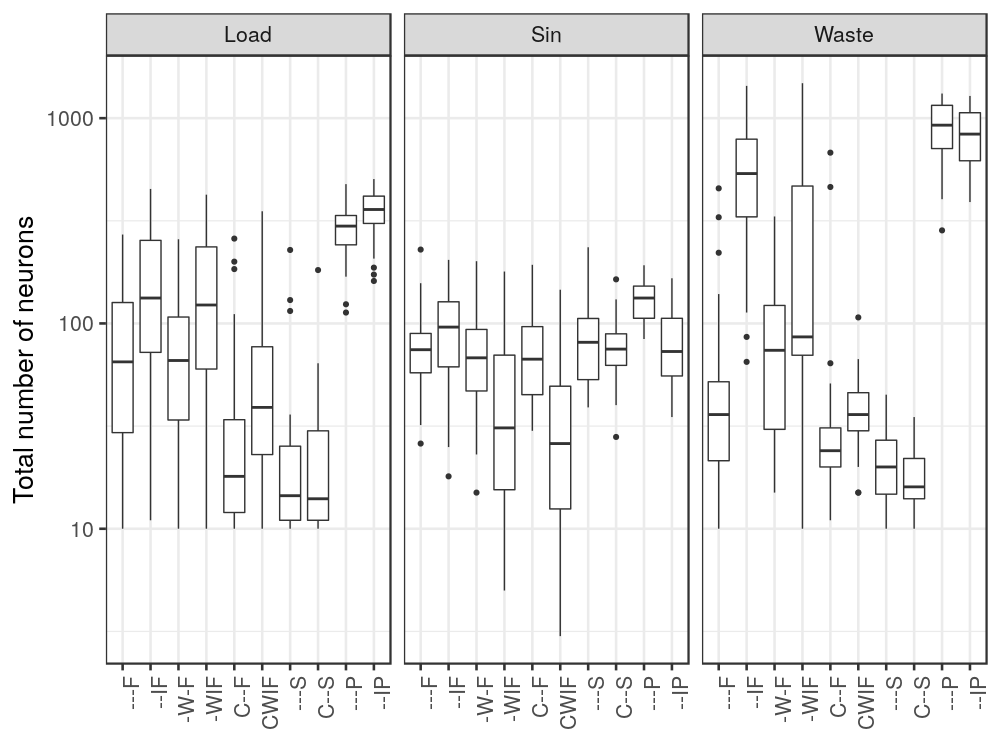}
  \caption{Distribution of the total number of LSTM cells}
  \label{figure:arch-nn}
\end{figure}

To get insights into the RNN architectures, we analyzed the (best) solutions. Figure~\ref{figure:arch-hl} shows the percentage of solutions that have a specific number of hidden layers (within the search space defined in Table~\ref{table:search-spaces}). Figure~\ref{figure:arch-lb} presents the percentage of solutions that have each of the possible look backs. Figure~\ref{figure:arch-nn} depicts the distribution of the total number of LSTM cells.

It is no surprise that the \emph{plain} encoding produced deeper and bigger (in terms of the total number of neurons) solutions, because of its own encoding limitations. On the other hand, two relatively similar combinations in terms of the error, namely \texttt{C--F} and \texttt{C--S}, present different architecture combinations. 

Also, it is quite interesting that there is no clear \emph{architecture trend}. There are some value ranges that seem to be more suitable, e.g, shallower instead of deeper networks, or mid-to-upper look back values for the load forecast problem, but we can not conclude that there is an \emph{all-rounder} architecture.

\subsection{Time Analysis}

The results presented in this study (Table~\ref{table:waste-summary}) show that using MRS as a proxy of the performance is as good as using short training results. However, as it is claimed in~\cite{Camero2018lowcost}, MRS is supposed to be a low-cost approach. Therefore, we compared the run time of Adam against MRS. Specifically, we randomly select 16 runs from the previous experiments (i.e., 100 architectures evaluated in 16 runs, totaling 1600 RNNs). Then, for each network we performed a MRS (100 samples) and a 10 epochs training using Adam.

We repeated the experiments because of two reasons. First, the previous experiments were run on a cluster of heterogeneous computers (hence the run times were not fairly comparable). Secondly, the final solutions were trained for 1000 epochs, thus the comparison would not have been fair.

\begin{figure}[ht]
  \centering
  \includegraphics[width=0.8\columnwidth]{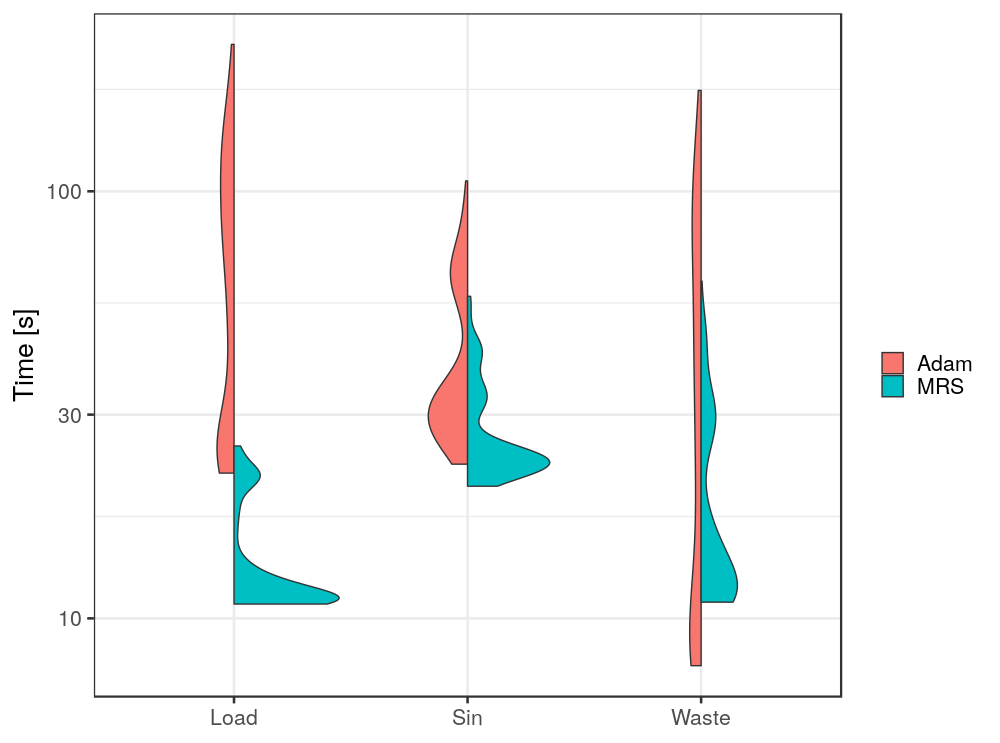}
  \caption{Time comparison: Adam (10 epochs) vs MRS (100 samples)}
  \label{figure:time-violin}
\end{figure}

Table~\ref{table:time-comparison} summarizes the time in seconds for both approaches, and Figure~\ref{figure:time-violin} shows the distribution of the time (in seconds). We performed a Wilcoxon rank sum test to compare both approaches. Note that we compare the overall results and the results of each problem independently. The results are presented in the table (\emph{Signif.}) using the following codes:  0 ‘***’ 0.001 ‘**’ 0.01 ‘*’ 0.05 ‘.’ 0.1 ‘ ’ 1.

On average, MRS is 2.6 times faster than Adam. These results are in line with the ones presented in~\cite{camero2019specialized}. In other words, if we have used the results of 10 epochs training using Adam to compare the architectures during the optimization process (instead of MRS), we will have spent more than twice the time!

\begin{table}
    \caption{Time comparison in seconds: Adam vs MRS. According to the Wilcoxon rank sum test, there is a significant improvement}
	\centering 
    \begin{tabular}{ llrrrr }
    \hline
    & [seconds] & Load	& Sin	& Waste & Overall \\
    \hline
    \multirow{5}{*}{Adam} 
    & Mean   &  72.1 &  41.8 &  45.3 &  53.1 \\
    & Median &  62.6 &  32.3 &  29.1 &  34.1 \\
    & Max    & 220.9 & 105.8 & 172.3 & 220.9 \\
    & Min    &  21.9 &  23.0 &   7.8 &   7.8 \\
    & Sd     &  48.7 &  19.0 &  40.8 &  40.5 \\
    \hline
    \multirow{5}{*}{MRS} 
    & Mean   &  13.8 &  27.9 &  19.3 &  20.3 \\
    & Median &  11.7 &  23.9 &  13.8 &  20.0 \\
    & Max    &  25.3 &  56.8 &  61.6 &  61.6 \\
    & Min    &  10.8 &  20.4 &  10.9 &  10.8 \\
    & Sd     &   4.3 &   8.0 &  10.7 &  10.0 \\
    \hline
    \multicolumn{2}{l}{Signif. (Adam vs MRS)} &  *** & *** & *** & *** \\
    \hline
	\end{tabular} \\    
    \label{table:time-comparison}
\end{table}

\subsection{Error Trade-off} 

Moreover, we studied how much the outcome of MRS is affected (i.e., error of the final solution) when the number of samples is changed. We repeated the \emph{waste} and \emph{load forecast} experiments using the \texttt{C--S} configuration, and 30, 50, and 200 samples per each solution evaluated (MRS).

Table~\ref{table:tradeoff} summarizes the error trade-off results. The Friedman rank sum test $p$-value is equal to 0.004996 (chi-squared = 12.84, df = 3) in the waste problem, while it is equal to 0.0003184 (chi-squared = 18.68, df = 3) in the load forecast problem. Therefore, we performed a pairwise comparison using the Conover test for a two-way balanced complete block design~\cite{conover1979multiple}, and the Holm $p$-value adjustment method. The results are presented in the row Conover of both tables. Groups sharing a letter are not significantly different ($\alpha=0.01$).

The results show that we might reduce the time (by taking fewer samples) but with an error increase. On the other hand, doubling the number of samples (used in this study), we will have not reduced the error. Nonetheless, it is quite interesting that even with a small number of samples, lets say 30, it is possible to estimate the performance of a network.

\begin{table}
    \caption{Waste and Load trade-off results. Groups sharing a letter in the Conover row are not significantly different}
	\centering 
    \begin{tabular}{ clrrrr }
    \hline
    & Samples	& 30	& 50	& 100	& 200 \\
    \hline
    \multirow{5}{1.5cm}{\centering Waste (MAE)}  
    & Mean	& 0.0734	& 0.0734	& 0.0725	& 0.0723 \\
    & Median	& 0.0732	& 0.0740	& 0.0723	& 0.0726 \\
    & Max	& 0.0778	& 0.0780	& 0.0760	& 0.0757 \\
    & Min	& 0.0694	& 0.0690	& 0.0688	& 0.0685 \\
    & Sd	& 0.0017	& 0.0020	& 0.0016	& 0.0018 \\

    \hline
    \multirow{5}{1.5cm}{\centering Load (MAPE)}  
    & Mean	    & 2.664	& 2.616	& 2.235	& 2.137 \\
    & Median	& 2.510	& 2.555	& 2.050	& 2.073 \\
    & Max   	& 4.436	& 3.750	& 4.605	& 3.146 \\
    & Min	    & 1.930	& 1.884	& 1.657	& 1.521 \\
    & Std	    & 0.597	& 0.492	& 0.564	& 0.405 \\

    \hline
    Conover & & a & a & b & b \\
    \hline
	\end{tabular} \\    
    \label{table:tradeoff}
\end{table}

\subsection{Algorithm Convergence}
\begin{figure}[ht]
  \centering
  \includegraphics[width=0.9\columnwidth]{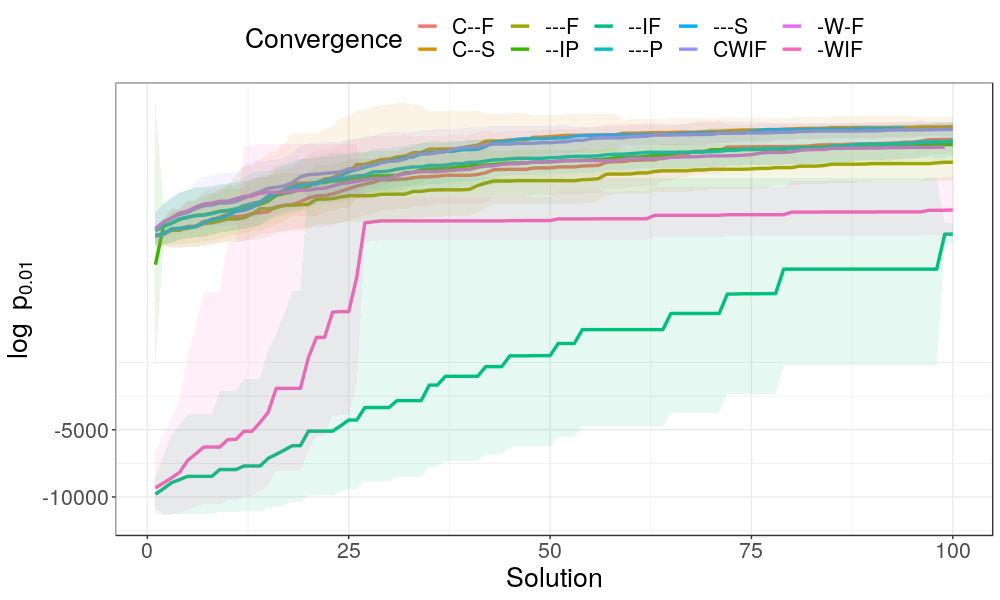}
  \caption{Average convergence of the fitness of the solutions for the waste problem}
  \label{figure:waste-convergence}
\end{figure}

\begin{figure}[ht]
  \centering
  \includegraphics[width=0.9\columnwidth]{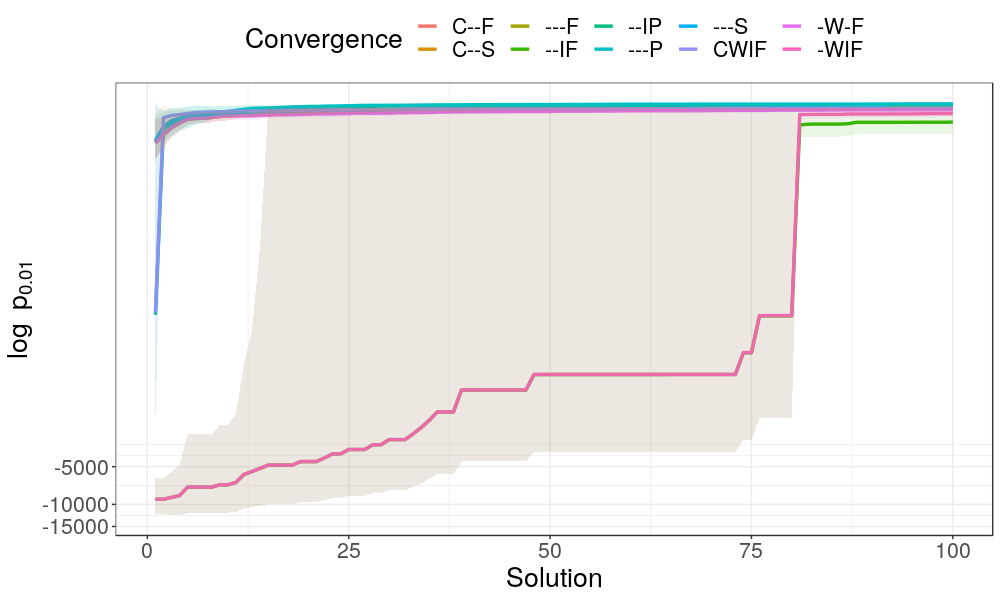}
  \caption{Average convergence of the fitness of the solutions for the load forecast problem}
  \label{figure:eunite-convergence}
\end{figure}

Finally, we studied the convergence of the proposed algorithm. Particularly, we analyzed the fitness (probability estimated by the MRS) of the solutions as the search was done. Figure~\ref{figure:waste-convergence} and~\ref{figure:eunite-convergence} depict the best-so-far MRS value against the number of candidates evaluated, average over all independent runs for each combination of encoding, warm-start, and constraint handling methods (shown by the bold line). 
Also, the standard deviation is illustrated by the shaded areas. It is important to point out that a higher MRS value is correlated with a better performance after training the network using Adam~\cite{Camero2018lowcost}, hence indicating that all combinations are converging.

Moreover, to show the impact of the penalty function, we compared the pairs \texttt{C--S} and \texttt{---S}, \texttt{C--F} and \texttt{---F}. Notice that the results present the average value of the MRS and the standard deviation (shaded area) for 30 independent runs (each combination) in the waste prediction problem. Therefore, we assume that the difference in performance (i.e., the convergence) can be explained by the the penalty.

\begin{figure}[!h]
  \centering
  \subfigure[]{
  	\includegraphics[width=0.45\textwidth]{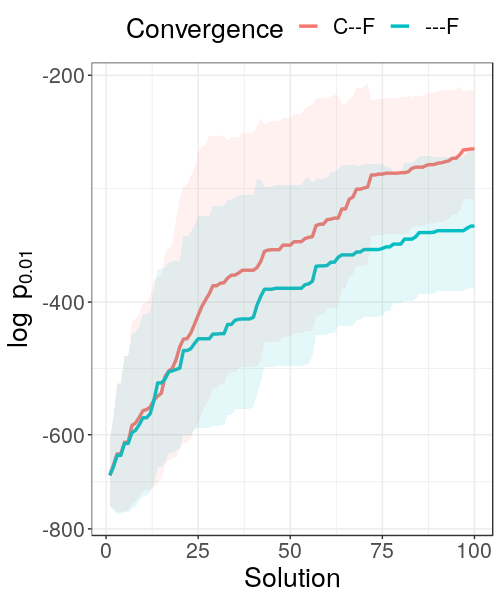}}
  \subfigure[]{
  	\includegraphics[width=0.45\textwidth]{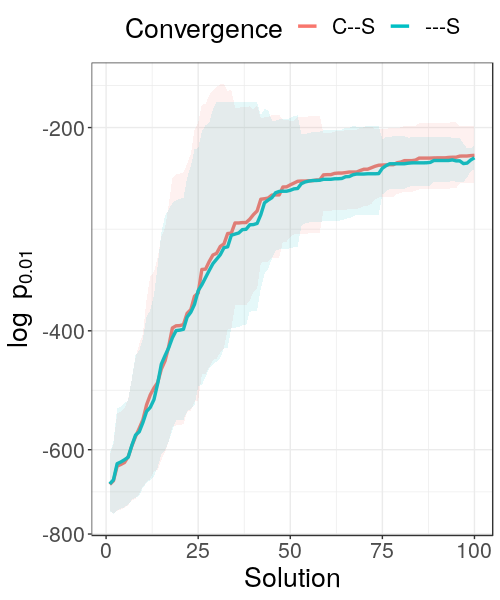}}
  \caption{Impact of the penalty on the average convergence (waste problem)}
  \label{figure:sine-logp-mae-1-2}
\end{figure}

\section{Conclusions and Future Work}
\label{sec:conclusions}

In this study, we propose to optimize the architecture of a recurrent neural network with a combination of Bayesian optimization and Mean Absolute Error Random Sampling (MRS). More specifically, we propose three fixed-length encoding schemes to represent variable size architectures (\emph{flag}, \emph{plain}, and \emph{size}), an alternative to deal with the many-to-one problem derived from the fixed-variable-length problem (i.e., the \emph{infeasible}solution), and two strategies to cope with the fixed-variable-length problem, namely \emph{warm-start} and \emph{constraints handling}.


We test our proposal on three prediction problems: the sine wave, the waste filling level of 217 bins in a metropolitan area of a city in Spain, and the maximum daily load forecast of an electricity company in Slovakia. We benchmark our proposal against state-of-the-art techniques, and we performed a time comparison and an error trade-off study. Notice that for each problem a different activation function has been used, namely, \texttt{tanh}, \texttt{sigmoid}, and \texttt{linear}.

The results show that none of the strategies presented outperforms the others in all cases. Nonetheless, using the \emph{size} encoding and the \emph{constraints handling} consistently show to be an \emph{effective} alternative to the problem.

Moreover, the results show that MRS is an \emph{efficient} alternative to optimize the architecture of an RNN. Particularly, we showed that evaluating an architecture using MRS is 2.6 times faster than performing a short training (ten epochs) using Adam, and without losing performance.

Overall, using BO, in combination with MRS, shows to be a competitive approach to optimize the architecture of an RNN. It offers a state-of-the-art error performance, while the time is drastically reduced.

Finally, for the next step, several issues have to be addressed. First, it is necessary to test on more data sets to validate the proposal. Second, MRS has to be further researched because it shows to be a promising alternative, but there is no clear explanation of why it works. Additionally, it will be interesting to use the \emph{warm start} strategy to explore \emph{augmenting restarts}, i.e., iteratively increase the number of hidden layers and feeding the model with the previous results.

\subsection*{Acknowledgments}

This work was supported in part by Universidad de M\'alaga, Andaluc\'ia Tech, Consejer\'ia de Econom\'ia y Conocimiento de la Junta de Andalu\'ia, Ministerio de Econom\'ia, Industria y Competitividad, Gobierno de Espa\~na, and European Regional Development Fund grant numbers TIN2017-88213-R (6city.lcc.uma.es), RTC-2017-6714-5 (ecoiot.lcc.uma.es), and UMA18-FEDERJA-003 (Precog). And by the Helmholtz Association’s Initiative and Networking Fund (INF) under the Helmholtz AI platform grant agreement (ID ZT-I-PF-5-1).

\balance

\bibliographystyle{elsarticle-harv}
\bibliography{bibliography}

\end{document}